\pdfoutput=1

\documentclass[11pt]{article}

\usepackage[final]{acl}

\usepackage{times}
\usepackage{latexsym}
\usepackage[T1]{fontenc}
\usepackage[utf8]{inputenc}
\usepackage{microtype}
\usepackage{inconsolata}
\usepackage{graphicx}
\usepackage{amsmath,amssymb}
\usepackage{algorithm}
\usepackage{algorithmic}
\usepackage{booktabs}
\usepackage{multirow}
\usepackage{url}
\usepackage{tikz}
\usetikzlibrary{arrows.meta,positioning,shapes.geometric,calc}
\title{What Drives Recovery in Agentic Text-to-Cypher?\\
LAST-CQ: An LLM Agent Self-Refinement
Framework}

\author{
  \textbf{Ioannis Prokopiou\textsuperscript{1,2}}\thanks{\ \,Equal contribution.},
  \textbf{Athanasios Aidinis\textsuperscript{2}}\footnotemark[1],
  \textbf{Panagiotis-Christos Kyrmpatsos\textsuperscript{2}},
  \textbf{Pantelis Vikatos\textsuperscript{2}}
\\
\\
  \textsuperscript{1}Athens University of Economics and Business, \quad
  \textsuperscript{2}Orfium
\\
  \small{
    \texttt{gian.prokopiou@aueb.gr}, \texttt{thanos.aidinis@orfium.com}
  } \\
  \small{
    \texttt{panagiotis.kyrmpatsos@orfium.com}, \texttt{pantelis@orfium.com}
  }
}

\begin{document}
\maketitle

% =============================================================================
\begin{abstract}
Agentic pipelines for structured-query generation are rapidly expanding, but it is unclear \emph{which} part of the loop produces the gain. We use LAST-CQ---a five-agent, training-free, execution-grounded Text-to-Cypher framework---as an instrumented testbed, running three counterfactuals over 2{,}471 live-database queries and six backbones spanning three vendor scale tiers. Removing correction is worth between $3.1\%$ aggregate execution-BLEU against the single-pass system and $12.3\%$ against a no-refinement counterfactual (up to $80.7\%$ for the weakest backbone). Replacing schema-grounded, LLM-synthesised feedback with raw database error strings costs almost nothing ($20.9\%$ vs.\ $19.9\%$ naive exact match; $<\!0.2\%$ end-to-end; equivalent within $\pm0.075$ set-F1 by two one-sided tests). Spending the same call budget on parallel sampling \emph{degrades} quality by $10$--$11\%$. What works is detecting failure and routing it to a retry, not the feedback sophistication or number of samples. LAST-CQ itself recovers $91.7\%$ of queries that fail under single-pass generation, while a query that succeeds first time still costs exactly one LLM call. We also show that n-gram overlap on serialised results is not a bound in either direction: it over-scores against set equivalence on $65.9\%$ of results while under-scoring against judged semantics. Finally, we calibrate our LLM judge against blind human labels and find it optimistic by $9$ points.
\end{abstract}

% =============================================================================
\section{Introduction}
\label{sec:intro}

Knowledge graphs underpin applications from biomedicine to financial compliance, and are queried through languages such as Cypher \cite{hogan2021knowledge,ji2022survey,francis2018cypher}. Making them reachable in natural language---the \emph{Text-to-Cypher} task---is what turns a graph database into something an agent can use as a tool \cite{ozsoy2025text2cypher,liu2026nli4db}. Large language models are capable of this in principle \cite{guo2023gpt4graph,rajkumar2022evaluating}, yet they routinely emit structurally hallucinated queries: non-existent labels, invalid relationship types, patterns that execute but return nothing \cite{xu2024hallucination}. In the graph setting this is particularly corrosive, because a well-formed but wrong query succeeds silently and returns misleading data.

The field's answer has been to wrap generation in an agentic loop: validate, execute, observe the failure, feed it back, regenerate \cite{madaan2023selfrefine,shinn2023reflexion,gusarov2025multiagent}. A loop bundles at least three mechanisms---failure detection, feedback synthesis, and re-generation---and reported gains are attributed to the architecture as a whole. If the gain in fact comes from the cheapest component, the field is over-investing in the expensive ones.

We treat this as the research question and build the instrument to answer it. \textbf{LAST-CQ} is a five-agent, training-free refinement framework for Text-to-Cypher: deterministic schema parsing at zero LLM calls, \texttt{EXPLAIN}-based pre-execution validation, schema-grounded correction, and empty-result relaxation. It recovers $91.7\%$ of queries that fail under single-pass generation, but its value here is that every mechanism in the loop can be removed independently and the loss measured.

Three counterfactuals over 2{,}471 queries and six backbones give a consistent answer. Removing correction entirely costs between $3.1\%$ against the single-pass system and $12.3\%$ against a no-refinement counterfactual, and up to $80.7\%$ for the weakest model. Removing \emph{schema grounding} from the feedback---keeping the detection and routing machinery, but replacing synthesised hints with the raw Neo4j \cite{neo4j} error string---costs $1.0$ point of exact match and under $0.2\%$ end-to-end. Spending the identical budget on parallel sampling makes things actively worse.

Our contributions:
\begin{enumerate}\itemsep2pt
    \item \textbf{An ablation of the agentic-refinement family, not another member.} Three controlled counterfactuals localise the gain in execution-grounded Text-to-Cypher loops to failure detection and routing, rather than to feedback sophistication or generation volume (\S\ref{sec:rq2}).
    \item \textbf{LAST-CQ, an instrumented testbed} whose components are individually removable, with full recovery and cost accounting across six backbones at three parameter scales (\S\ref{sec:system}, \S\ref{sec:rq1}).
    \item \textbf{Order-invariant evaluation of agentic structured-query generation.} We report judge-based semantic accuracy over 13{,}563 executed results for all six backbones, set-based precision/recall/F1/Jaccard against the retry baseline, and a quantification of how n-gram overlap diverges from both stricter and looser notions of correctness (\S\ref{sec:rq3}).
\end{enumerate}

% =============================================================================
\section{Related Work}
\label{sec:related}

\subsection{Execution-Grounded Refinement}

Mapping language to executable form has been studied most intensively as Text-to-SQL \cite{yu2018spider,kamath2019survey,zettlemoyer2005learning,dong2016language} and especially with execution-guided agents \cite{deng2025reforce,zhang2025exesql}. Modern pipelines are multi-stage: schema linking \cite{lei2020reexamining,wang2020ratsql}, LLM generation \cite{li2023bird,pourreza2023dinsql}, and execution-guided refinement against a live executor \cite{wang2018execguided,zhai2025execfeedback,ni2023lever}. Self-Refine \cite{madaan2023selfrefine} and Reflexion \cite{shinn2023reflexion} show models improve their own outputs from retrospective feedback; recent work adds prospective reflection \cite{wang2026preflect} and dynamic agent topologies \cite{lu2026dytopo,wu2024autogen,acharya2025agentic}. The shared premise is that richer feedback yields better correction. \S\ref{sec:rq2} tests that premise and finds it largely unsupported here.

\subsection{Text to Cypher, SPARQL and Graph RAG}

Techniques transfer to the property-graph setting with adaptations \cite{zhao2023cyspider,tran2024robust,guo2022spcql,opitz2022zerohero,zhao2023s2ctrans}. \citet{ozsoy2025text2cypher} release the benchmark we use, \citet{ozsoy2025schemafilter} cut its token cost by schema filtering, and \citet{tiwari2025autocypher} generate training data verified by execution. Commercial tooling such as Neo4j's NeoDash \cite{neo4j_neodash} retries failed queries on raw database errors, without published benchmark evaluation. Graph RAG methods \cite{edge2024graphrag,baek2023kaping,sun2024thinkongraph,he2024gretriever,ma2025llmkgqa} bypass query formulation by retrieving subgraph context, trading precision and index utilisation for robustness; the Text-to-SPARQL literature faces the same grounding problems on RDF \cite{banerjee2022modern,kovriguina2023sparqlgen,omar2023universal}.

The closest published system is \citet{gusarov2025multiagent}, and the architectural overlap is substantial: both are multi-agent Text-to-Cypher pipelines built on schema grounding, execution, failure detection, feedback generation and iterative correction. They use Memgraph \cite{memgraph} and eight modules, we Neo4j and five agents; their schema is formatted into the generator's prompt, ours parsed deterministically at zero LLM calls; their correction is entirely post-execution, ours adds \texttt{EXPLAIN} validation \emph{before} execution so malformed queries never reach the database; they re-ground hallucinated entities by fuzzy string similarity plus LLM semantic ranking over database contents, we by schema-grounded hints and a fixed relaxation set; they cap correction at four iterations, we use two validation cycles and one relaxation pass. Evaluation differs most: they report LLM-judge accuracy on 750 CypherBench questions across four backbones, with no ablation and no recovery analysis; we report execution-grounded, order-invariant and judge-based metrics on 2{,}471 questions across six backbones, with per-component ablations and two counterfactual baselines.

We do not claim the architectural family is novel. \citet{gusarov2025multiagent} and this work are plainly members of it, which is precisely the motivation. When several groups converge independently on the same loop, the question stops being whether to build one and becomes which part is load-bearing. \S\ref{sec:rq2} answers it for this family, and \S\ref{sec:naive} compares our answer with the one prior system to ask a version of the same question \cite{tomczak2026cygnet}. Whether the answer transfers to a Memgraph pipeline with fuzzy entity re-grounding is an empirical question we do not settle here, since we run no experiment on their engine or benchmark; we claim only that the counterfactual design is directly reusable on it.

% =============================================================================
\section{LAST-CQ: An Instrumented Testbed}
\label{sec:system}

\subsection{Agents}
\label{sec:agents}

\begin{figure*}[htbp]
\centering
\includegraphics[width=\linewidth]{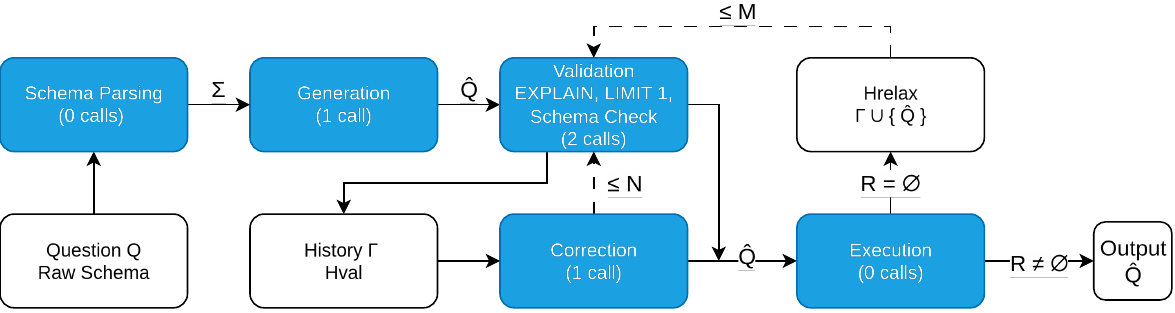}
\caption{Architecture of LAST-CQ. Candidate Cypher queries are iteratively refined using validation and execution feedback to narrow the search space under schema constraints. Colored boxes represent agents, and LLM calls are shown in parentheses.}
\label{fig:system_architecture}
\end{figure*}

LAST-CQ decomposes generation into five agents orchestrated as a state machine (Figure~\ref{fig:system_architecture}). We state each agent's LLM cost explicitly, since \S\ref{sec:rq2} turns on which costs buy anything.

\noindent\textbf{Schema Parsing} $\mathcal{P}$ (0 LLM calls) converts a raw schema into a normalised representation of nodes, properties and relationships using deterministic parsers covering eight formats, with LLM extraction only as fallback; on this benchmark the deterministic path handles all 2{,}471 schemas and the fallback never fires.

\noindent\textbf{Generation} $G$ (1 call) builds a candidate from the parsed schema and question, following \citet{ozsoy2025text2cypher}.

\noindent\textbf{Validation} $\mathcal{V}$ (0--2 calls) runs \texttt{EXPLAIN} against the live schema, then executes with \texttt{LIMIT~1} to catch queries that are well-formed but return nothing. If both checks pass it returns immediately and \emph{invokes no model at all}. Only on failure does it parse the query (1 call) and run a deterministic schema check---do these labels, relationship types and properties exist?---and only if that check reports violations does a second call synthesise the grounded hints $H_{\text{val}}$.

\noindent\textbf{Correction} $C$ (1 call) rewrites the query from the current hints, the schema $\Sigma$, and the correction history $\Gamma$, which is passed in the prompt with an instruction not to repeat any prior attempt.

\noindent\textbf{Execution} $\mathcal{E}$ (0 calls) runs the validated query and, if it returns nothing, injects a \emph{fixed} constant $H_{\text{relax}}$: six hints covering case-insensitive regex, substring rather than equality, canonical \texttt{datetime()} formatting, dropping over-restrictive labels and simplifying \texttt{RETURN}. It then routes back to Correction. No model produces $H_{\text{relax}}$; the rewriting is done by the Correction Agent.

\subsection{Formal Framework}

Given a question $Q$ and schema $\Sigma$, the objective is a query $\hat{Q}\in\mathcal{L}_{\text{Cypher}}$ consistent with $\Sigma$ that captures the user's intent. Execution returns a non-empty result set $R$, $\emptyset$, or a failure $\mathcal{E}_{\text{fail}}$; following common practice we adopt the criterion $\mathrm{Correct}(\hat{Q}) \iff \mathcal{E}(\hat{Q}) = R \neq \emptyset$, while acknowledging that a non-empty result does not imply semantic correctness (\S\ref{sec:rq3}).

The refinement state is $(\hat{Q}, H, \Gamma, \Sigma)$, with $\Gamma=(\hat{Q}_1,\ldots,\hat{Q}_{t-1})$ the history of attempted queries. The agents are $\mathcal{P}(\Sigma)\!\rightarrow\!\Sigma$, $G(Q,\Sigma)\!\rightarrow\!\hat{Q}$, $\mathcal{V}(\hat{Q},\Sigma)\!\rightarrow\!(\textsf{valid},H_{\text{val}})$, $C(\hat{Q},H,\Gamma,\Sigma)\!\rightarrow\!\hat{Q}$ and $\mathcal{E}(\hat{Q})\!\rightarrow\! R\cup\{\emptyset,\mathcal{E}_{\text{fail}}\}$. The two hint sources differ in kind, since $H_{\text{val}}$ is LLM-synthesised from a deterministic schema analysis while $H_{\text{relax}}$ is a hard-coded constant, and \S\ref{sec:rq2} explains that difference.

\paragraph{Control flow and termination.} In Algorithm~\ref{alg:pipeline} the inner loop alternates Validation and Correction until the query is valid or $N$ iterations are spent; the outer loop executes it and, on an empty result, injects $H_{\text{relax}}$ and refines once more, for up to $M$ cycles. It terminates when the query validates and returns $R\neq\emptyset$, when the validation budget is spent, or when the relaxation pass has already fired, returning the result whether or not it is empty.

\paragraph{What a query actually costs.} Because Validation short-circuits with no model call when both checks pass, a query that succeeds first time costs exactly one LLM call, the same as single-pass generation, while each failed cycle adds up to three (parse, hint synthesis, correction). Since $53$--$97\%$ of queries need no correction (Appendix~\ref{app:extra}), the overhead falls on the minority that would otherwise return nothing. Table~\ref{tab:main} reports mean pipeline \emph{stages}, which our traces record directly; derived LLM-call means range from about $1.1$ (Claude) to $2.5$ (Gemini Flash).

\begin{table}[t]
\centering\small
\begin{tabular}{ll}
\toprule
\textbf{Symbol} & \textbf{Description} \\
\midrule
$Q$ & Natural language question \\
$\hat{Q}$ & Current Cypher query \\
$\Sigma$ & Schema, after parsing \\
$H_{\text{val}}$ & Validation hints (LLM-synthesised) \\
$H_{\text{relax}}$ & Relaxation hints (constant) \\
$\Gamma$ & History of attempted queries \\
$R$, $\mathcal{E}_{\text{fail}}$ & Non-empty result; execution failure \\
$N,M$ & Validation / execution budgets \\
$\mathcal{P},G,\mathcal{V},C,\mathcal{E}$ & The five agents, in pipeline order \\
\bottomrule
\end{tabular}
\caption{Notation used throughout.}
\label{tab:notation}
\end{table}

\begin{algorithm}[t]
\caption{Agentic query generation with iterative validation and execution feedback}
\label{alg:pipeline}
\begin{algorithmic}[1]
\REQUIRE question $Q$, schema $\Sigma$, budgets $N$,$M$
\ENSURE Cypher query $\hat{Q}$

\STATE $\Sigma \leftarrow \mathcal{P}(\Sigma)$ \COMMENT{Normalize schema}
\STATE $\hat{Q} \leftarrow G(Q,\Sigma)$
\STATE $\Gamma \leftarrow \emptyset$

\FOR{$j=1$ to $M$}

    \FOR{$t=1$ to $N$}

        \STATE $(v,H) \leftarrow \mathcal{V}(\hat{Q},\Sigma)$

        \IF{$v=\textsf{valid}$}
            \STATE \textbf{break}
        \ENDIF

        \STATE $\Gamma \leftarrow \Gamma \cup \{\hat{Q}\}$
        \STATE $\hat{Q} \leftarrow C(\hat{Q},H,\Gamma,\Sigma)$

    \ENDFOR

    \STATE $R \leftarrow \mathcal{E}(\hat{Q})$

    \IF{$R\neq\emptyset$}
        \STATE \textbf{return} $\hat{Q}$
    \ENDIF

    \STATE $H \leftarrow$ execution-derived hints
    \STATE $\Gamma \leftarrow \Gamma \cup \{\hat{Q}\}$
    \STATE $\hat{Q} \leftarrow C(\hat{Q},H,\Gamma,\Sigma)$

\ENDFOR

\STATE \textbf{return} $\hat{Q}$
\end{algorithmic}
\end{algorithm}

\subsection{Worked Example}
\label{sec:example}

Figure~\ref{fig:trace} shows two real traces, one per hint type. In the first, the generated query calls \texttt{YEAR()}, a SQL function absent from Cypher; \texttt{EXPLAIN} rejects it, the deterministic schema check confirms no such function, and the synthesised hint names the Cypher equivalent, which the Correction Agent applies. In the second, the query is syntactically valid and passes \texttt{EXPLAIN}, but returns nothing because the filter is case- and substring-sensitive; the Execution Agent injects $H_{\text{relax}}$, and the corrected query adopts the case-insensitive regex form named literally in hint three of that constant set. The contrast matters for \S\ref{sec:rq2}: the first correction required a model to look at the schema, the second required only a fixed string.

\begin{figure}[t]
\centering\footnotesize
\begin{tabular}{@{}p{0.94\columnwidth}@{}}
\toprule
\textbf{(a) Validation hint (schema-grounded, 1 LLM call).} \\
$Q$: \emph{Average star rating of businesses reviewed by users in each year?} \\
$\hat{Q}_0$: \texttt{... WITH YEAR(r.date) AS year ...} \\
$\mathcal{V}$: \texttt{EXPLAIN} fails---\texttt{YEAR()} is not a Cypher function \\
$\hat{Q}_1$: \texttt{... WITH date(r.date).year AS year ...} \\
\midrule
\textbf{(b) Relaxation hint (constant, 0 LLM calls).} \\
$Q$: \emph{List the first 3 articles that mention new energy technologies.} \\
$\hat{Q}_0$: \texttt{... WHERE org.name CONTAINS "new energy} \\ \texttt{\ \ technologies" ...} \\
$\mathcal{E}$: valid, but $\mathcal{R}=\emptyset$ $\Rightarrow$ inject $H_{\text{relax}}$ \\
$\hat{Q}_1$: \texttt{... WHERE org.name =\textasciitilde\ '(?i).*new} \\ \texttt{\ \ energy technologies.*' ...} \\
\bottomrule
\end{tabular}
\caption{Two real benchmark traces. (b)'s rewrite applies hint three of the constant set $H_{\text{relax}}$ verbatim.}
\label{fig:trace}
\end{figure}

% =============================================================================
\section{Experimental Setup}
\label{sec:setup}

\paragraph{Dataset and models.} We use the executable subset of the Neo4j Text2Cypher benchmark \cite{ozsoy2025text2cypher}, namely 2{,}471 question--query pairs across 16 domains with a live Neo4j demonstration database, so both generated and reference queries can actually be run. We use it unmodified, and it is the same subset as \citet{ozsoy2025schemafilter}. Six backbones span three scales: GPT-4o-mini and GPT-4o \cite{openai2024gpt4o}, Gemini 1.5 Flash and Pro \cite{gemini2024v15}, Claude Sonnet 4 \cite{anthropic2025claude4}, and DeepSeek-Chat, whose endpoint maps to DeepSeek-V3 \cite{deepseek2024v3}. All run at $T=0$. No fine-tuning is performed; every condition is in-context prompting (Appendix~\ref{app:models}).

\paragraph{Conditions.} \emph{Baseline} is single-pass generation that adopts the prompting methodology of \citet{ozsoy2025text2cypher} and uses none of their released fine-tuned checkpoints. Every model in every condition of this paper is an off-the-shelf instruction-tuned backbone prompted in context; no weights are updated anywhere. For the four backbones those authors also evaluate (GPT-4o, GPT-4o-mini, Gemini 1.5 Flash, Gemini 1.5 Pro) we take their reported \emph{prompting-only} figures on this same executable subset, and for the two they do not cover (Claude Sonnet 4, DeepSeek-Chat) we ran their prompting method ourselves under identical conditions.

We are not using their fine-tuned models as a comparison point because fine-tuning gains roughly $0.13$--$0.34$ Google-BLEU on their metrics---more than any inference-time result here, and we do not claim to match it---but it needs labelled data and one training run per backbone. This paper asks an orthogonal question: given a frozen model behind an API, which part of a refinement loop earns its keep.

\emph{LAST-CQ} is the pipeline with schema access only. It parses, normalises and validates statically, but cannot execute queries. We set $N=2$ from the observed cycle distribution and $M=1$; the implementation realises $M=1$ as a single-shot latch rather than a tunable counter, so larger $M$ was never measured. \emph{LAST-CQ+DB} adds the live database, $N=2$ validation cycles and $M=1$ relaxation pass. \emph{Naive retry} and \emph{Best-of-3} are the counterfactual baselines of \S\ref{sec:rq2}.

\emph{No-refinement} is not a separate run: it re-scores the \emph{same} LAST-CQ+DB execution with every query that entered the correction loop assigned zero, i.e.\ treated as the unrecoverable failure it would have been in a non-agentic pipeline. Baseline and no-refinement are thus distinct in kind---one is an independently executed system, the other a counterfactual upper bound---and we report both.

\paragraph{Metrics and their invariances.} Each metric is labelled by what it is invariant to, since \S\ref{sec:rq3} shows they disagree. \emph{Execution Google-BLEU} \cite{wu2016googlenmt} computes $\frac{1}{4}\sum_{n=1}^{4}\min(\text{prec}_n,\text{rec}_n)$ between the \emph{serialised database results} of the generated and reference queries; it is order-sensitive and we treat it as a comparability metric, not as evidence of correctness. \emph{Execution Exact Match} compares result sets after canonical normalisation that sorts rows and keys, so it is order-invariant by construction. \emph{Judge semantic accuracy} rates each executed result CORRECT, PARTIAL or INCORRECT and is likewise order-invariant. Both are reported on the full corpus for all six models. \emph{Set-based precision, recall, F1, Jaccard and set equality} treat each payload as a set of canonicalised rows and are order-invariant. We also report \emph{hit retrieval} and \emph{recovery rate}. Definitions are in Appendix~\ref{app:extra}, and the judge protocol in Appendix~\ref{app:judge}.

One limitation is that per-query result payloads were retained for the 1{,}917-query correction subset across all six models and for a 214-query paired subset on two models, but not for the full corpus, so set-based numbers are reported on those subsets while full-corpus order-invariant evidence rests on exact match and the judge. Two conventions matter. The benchmark's GLEU convention credits a non-empty result against an \emph{empty} reference with $1.0$, whereas set metrics do not; we use the strict set-theoretic convention for all set-based numbers and quantify the effect in \S\ref{sec:naive}. And ``aggregate'' means the unweighted mean over the six backbones, not a pooling over queries. The two differ because failure counts are unevenly distributed, so where a pooled figure is more informative we say so (e.g.\ the $93.8\%$ pooled recovery rate against the $91.7\%$ per-model mean).

% =============================================================================
\section{RQ1: Does the Framework Work?}
\label{sec:rq1}

\subsection{Execution Quality}

\begin{table*}[t]
\centering\small
\begin{tabular}{lcccc|ccccc}
\toprule
& \multicolumn{3}{c}{\textbf{Execution GLEU}} & \textbf{Sem.\ Acc.}
& \multicolumn{5}{c}{\textbf{Recovery of single-pass failures}} \\
\cmidrule(lr){2-4}\cmidrule(lr){5-5}\cmidrule(lr){6-10}
\textbf{Model} & \textbf{Base} & \textbf{LAST-CQ} & \textbf{+DB} & \textbf{+DB (\%)}
& \textbf{Fail} & \textbf{Rec.} & \textbf{Rate} & \textbf{$\Delta$HR} & \textbf{Stages} \\
\midrule
GPT-4o-mini$^{\dagger}$ & 0.4180 & \textbf{0.4431} & 0.4378 & 85.4 & 306  & 254  & 0.83 & $+0.10$ & 3.54 \\
Gemini-Flash-1.5        & 0.4018 & 0.4056 & \textbf{0.4350} & 85.0 & 1128 & 1098 & 0.97 & $+0.44$ & 4.19 \\
\midrule
Claude Sonnet 4         & 0.4743 & 0.4732 & \textbf{0.4821} & 91.0 & 63   & 61   & 0.97 & $+0.02$ & 3.14 \\
Gemini-Pro-1.5          & 0.4100 & 0.4155 & \textbf{0.4258} & 86.1 & 130  & 106  & 0.82 & $+0.04$ & 3.30 \\
\midrule
GPT-4o                  & 0.5270 & 0.5139 & \textbf{0.5421} & 89.6 & 206  & 202  & 0.98 & $+0.08$ & 4.18 \\
DeepSeek-Chat           & \textbf{0.5212} & 0.5155 & 0.5146 & 91.6 & 84   & 78   & 0.93 & $+0.03$ & 3.20 \\
\midrule
\textit{Aggregate}      & \textit{0.4587} & \textit{0.4611} & \textit{0.4729} & \textit{88.2} & \textit{1917} & \textit{1799} & \textit{0.917} & \textit{$+0.118$} & \textit{3.59} \\
\bottomrule
\end{tabular}
\caption{Left: execution-based Google-BLEU between database results of generated and reference queries, plus order-invariant judge semantic accuracy (CORRECT $+$ PARTIAL) over 13{,}563 LAST-CQ+DB results. Right: recovery of queries that fail under single-pass generation, and its cost in mean pipeline stages per query (single-pass $=1$). Stages are agent invocations, not all of which call a model: Schema Parsing and Execution never do, and Validation does not when the query passes (\S\ref{sec:system}). Aggregate is the unweighted mean over the six models, except Fail/Rec.\ which are counts; the pooled recovery rate is $1{,}799/1{,}917 = 93.8\%$ against the $91.7\%$ per-model mean. The judge was run on LAST-CQ+DB outputs only. $^{\dagger}$GPT-4o-mini is also the judge model.}
\label{tab:main}
\end{table*}

Table~\ref{tab:main} reports execution-grounded quality. LAST-CQ+DB improves average GLEU by $3.4\%$ over Baseline ($+6.5\%$ small, $+2.8\%$ medium, $+2.9\%$ for GPT-4o). Schema-only LAST-CQ yields about $+1\%$ and slightly \emph{decreases} GLEU for the two strongest models: static validation without execution evidence triggers unnecessary rewrites when baseline quality is already high. DeepSeek-Chat decreases by $1.3\%$, which Appendix~\ref{app:extra} traces to schema-representation differences rather than correction drift. Exact match improves more modestly ($+0.5\%$), consistent with the loop's job of recovering failures.

Semantic accuracy is high and does not track GLEU rank order: DeepSeek has the lowest GLEU gain and the highest judged accuracy. It describes LAST-CQ+DB's output quality rather than showing the loop \emph{improves} semantics, because the judge was not run on baseline outputs, whose payloads were not retained. We therefore cannot establish from the judge that refinement raises semantic correctness relative to single-pass generation, only that the refined outputs are largely valid; the paired, order-invariant evidence in \S\ref{sec:naive} is what compares conditions directly. \S\ref{sec:rq3} calibrates this column against a second judge and against human labels, and finds it optimistic; it should be read accordingly.

\subsection{Recovery and Cost}

The right half of Table~\ref{tab:main} is the system's strongest result. In a non-agentic pipeline a failed query yields an error or an empty result with no recovery path; LAST-CQ+DB recovers $91.7\%$ of them, lifting hit retrieval by $11.8\%$, at $3.14$--$4.19$ pipeline stages per query, most of which invoke no model (\S\ref{sec:system}). Gemini Flash is the informative outlier: half its queries fail at baseline, $97.3\%$ are recovered, and its mean recovered GLEU of $0.44$ matches strong models' first-pass output. Behaviour is stable across the 16 domains (CV $=0.147$), and the deterministic parser handles $100\%$ of schemas at zero LLM calls. Given a framework that works, the question this paper exists to answer is which part of it is doing the work.

% =============================================================================
\section{RQ2: What Actually Drives Recovery?}
\label{sec:rq2}

We remove one candidate mechanism at a time: correction itself, the schema grounding of the feedback, and the sequential-correction strategy.

\subsection{Removing Correction Entirely}

\begin{table}[t]
\centering\small
\begin{tabular}{lcccl}
\toprule
\textbf{Model} & \textbf{Base} & \textbf{Full} & \textbf{No-Ref.} & \textbf{$\Delta$} \\
\midrule
GPT-4o-mini & 0.4180 & 0.4378 & 0.4069 & $+7.6\%$ \\
Gem.-Flash  & 0.4018 & 0.4350 & 0.2407 & $+80.7\%$ \\
\midrule
Claude S4   & 0.4743 & 0.4821 & 0.4708 & $+2.4\%$ \\
Gem.-Pro    & 0.4100 & 0.4258 & 0.4096 & $+3.9\%$ \\
\midrule
GPT-4o      & 0.5270 & 0.5421 & 0.4977 & $+8.9\%$ \\
DeepSeek    & 0.5212 & 0.5146 & 0.5011 & $+2.7\%$ \\
\midrule
\textit{Aggregate} & \textit{0.4587} & \textit{0.4729} & \textit{0.4211} & \textit{$+12.3\%$} \\
\bottomrule
\end{tabular}
\caption{Removing correction, bounded two ways. Base: the independent single-pass system. No-Ref.: the \emph{same} LAST-CQ+DB run re-scored with corrected queries zeroed. $\Delta$ is Full over No-Ref.; all $p<0.001$ (paired bootstrap, $n=10{,}000$).}
\label{tab:ablation}
\end{table}

Table~\ref{tab:ablation} removes correction, and we bound the effect from both sides rather than quoting the flattering number. Against the \emph{no-refinement counterfactual}---the same run re-scored with corrected queries zeroed---the gain is $12.3\%$ in aggregate and $80.7\%$ for Gemini Flash, significant for every model ($p<0.001$) including the strongest, where under $4\%$ of queries enter the loop. But that assumes a failed query is worth nothing, which upper-bounds correction's contribution. Against the \emph{deployed alternative}, single-pass generation, the gain is $0.4729$ vs.\ $0.4587$, or $+3.1\%$. The true value lies between these bounds---nearer the lower for strong backbones, the upper for weak ones. Either way, \emph{some} correction is worth having. Which one is the question we turn to next.

\subsection{Removing Schema Grounding}
\label{sec:naive}

\begin{table}[t]
\centering\small
\resizebox{\columnwidth}{!}{%
\begin{tabular}{lrrrrrr}
\toprule
& \multicolumn{2}{c}{\textbf{EM (\%)}} & \multicolumn{2}{c}{\textbf{GLEU}} & \multicolumn{2}{c}{\textbf{Naive, set}} \\
\cmidrule(lr){2-3}\cmidrule(lr){4-5}\cmidrule(lr){6-7}
\textbf{Model} & \textbf{Naive} & \textbf{L-CQ} & \textbf{Naive} & \textbf{L-CQ} & \textbf{F1} & \textbf{Acc.} \\
\midrule
GPT-4o-mini & 13.7 & 13.7 & 0.248 & 0.249 & 0.145 & 14.4 \\
Gem.-Flash$^{\ddagger}$ & 18.7 & 19.9 & 0.412 & 0.426 & 0.221 & 21.7 \\
\midrule
Claude S4   & 25.4 & 27.0 & 0.431 & 0.444 & 0.270 & 27.0 \\
Gem.-Pro$^{\ddagger}$ & 15.4 & 16.9 & 0.285 & 0.307 & 0.198 & 19.2 \\
\midrule
GPT-4o      & 36.4 & 38.8 & 0.527 & 0.533 & 0.393 & 39.3 \\
DeepSeek    & 20.2 & 17.9 & 0.455 & 0.396 & 0.250 & 25.0 \\
\midrule
\textit{Aggr.} & \textit{19.9} & \textit{20.9} & \textit{0.392} & \textit{0.400} & \textit{0.229} & \textit{22.6} \\
\bottomrule
\end{tabular}}%
\caption{Naive retry vs.\ LAST-CQ on the 1{,}917 corrected queries. EM is the published BLEU$\,\geq\,$0.999 proxy; ``set'' columns give true order-invariant F1 and set equality for the naive arm. $^{\ddagger}$Both arms use Gemini~2.5~Flash for the retired 1.5 models; comparisons are within-model.}
\label{tab:naive}
\end{table}

Naive retry is not feedback-free: it keeps LAST-CQ's entire detection and routing machinery---\texttt{EXPLAIN} validation, \texttt{LIMIT~1} empty detection, the cycle budget, the history $\Gamma$, the routing back into re-validation. It removes only the \emph{content} of $H_{\text{val}}$: the Correction Agent receives the raw Neo4j error string instead of LLM-synthesised grounded guidance, mirroring what deployed tooling such as NeoDash \cite{neo4j_neodash} does. We re-ran all 1{,}917 corrected queries across six models at the same budget $N=2$.

\paragraph{Why 1{,}917 and not 2{,}471.} The 1{,}917 figure is a pooled count over the six backbones of the model--query pairs on which the Correction Agent actually fired ($306+1128+63+130+206+84$), not a subset of any single model's 2{,}471 questions. The counterfactual changes only the content of the hint handed to that agent, so on every pair a model answered on the first pass the correction path is never entered in either arm: both conditions emit the same query and the same result, and scoring them adds an identical constant to both without separating the treatments. Those pairs are not excluded from the headline comparison, since the end-to-end aggregate reported below is computed over the full corpus. One consequence should be stated plainly: because the set is defined by the queries LAST-CQ itself chose to correct, it measures relative \emph{correction} quality under a shared detector, not relative end-to-end system quality.

The grounding buys almost nothing (Table~\ref{tab:naive}). Naive retry reaches $19.9\%$ exact match against LAST-CQ's $20.9\%$, and end-to-end the aggregate GLEU gap is $0.4729$ vs.\ $0.4718$, under $0.2\%$. This is not an aggregation artifact: LAST-CQ leads on five of six models by one to two points and \emph{loses} on DeepSeek ($17.9\%$ vs.\ $20.2\%$). Excluding the two Gemini-substituted models leaves the picture unchanged (naive set-F1 $0.248$ on the remaining 659 queries against $0.229$ overall).

Order-invariant metrics confirm rather than overturn this. On the 214 queries for which both systems' result payloads were retained (Gemini-Pro and DeepSeek), set-based F1 is $0.200$ for LAST-CQ against $0.219$ for naive retry, a difference of $-0.019$. Because a null is not itself evidence of equivalence, we test it directly with two one-sided tests: the difference is equivalent to zero within $\pm 0.075$ set-F1 ($p = 0.002$) but not within $\pm 0.05$ ($p = 0.053$), so we can rule out a schema-grounding advantage larger than about $0.075$. Set equality and Jaccard agree, and restricting to the 148 rows where both pipelines serialise ground truth identically leaves it unchanged ($-0.023$).

\paragraph{The empty-result convention flatters both systems.} Of the 1{,}917 corrected queries, $230$ ($12.0\%$) have an \emph{empty} reference result, and the benchmark's GLEU convention awards $1.0$ whenever the reference is empty and the output is not. A query that should return nothing therefore scores perfectly for returning something, which is exactly the failure the relaxation pass would induce. Excluding them, the two systems separate the other way: GLEU $0.350$ for LAST-CQ against $0.365$ for naive retry on the remaining $1{,}687$, and set-F1 $0.070$ against $0.104$ on the paired subset (Appendix~\ref{app:extra}). LAST-CQ's small aggregate edge in Table~\ref{tab:naive} is carried entirely by the subset where the metric is most permissive. We report this because it cuts against our own system: once the forgiving cases go, schema-grounded feedback does not merely fail to assist; it provides no actionable guidance.

Independent evidence points the same way. \citet{tomczak2026cygnet}, working on a different benchmark (CypherBench), a different gate design and a different model set, ablates corrector feedback formats and finds a plain verbal rendering of the database error reaches nearly the correction success of the richest category-typed structured feedback, at a fraction of the input-token cost.

The conclusion is not that the framework is unnecessary, but that its value sits somewhere other than where the literature has been investing. Detection and orchestration---knowing a query failed, knowing which failure it was, routing it back on a bounded budget---produce the $91.7\%$ recovery and the ablation gap. The sophistication of the hint string, where the extra LLM call goes, produces about one point of exact match. That is the result for a field building ever more elaborate feedback-synthesis modules.

\subsection{Spending the Budget on Sampling}

The third counterfactual keeps the budget and changes how it is spent: three candidates sampled in parallel ($k{=}3$, $T{=}0.7$), the first non-empty result selected. Despite matching LAST-CQ+DB's budget, Best-of-3 \emph{underperforms even single-pass generation}, at GLEU $0.3728$ vs.\ $0.4180$ for GPT-4o-mini and $0.4671$ vs.\ $0.5270$ for GPT-4o, degradations of $10.8\%$ and $11.4\%$, against LAST-CQ+DB's $0.4378$ and $0.5421$. GPT-4o-mini's hit rate actually \emph{falls} against its own $T{=}0$ baseline ($87.6\%\!\rightarrow\!85.2\%$): on hard queries, higher temperature raises the chance that \emph{every} candidate hallucinates. In terms of diversity, only $1.37$ and $1.16$ candidates were tried on average, with $79$--$91\%$ of queries answered by the first.

Sampling requires $T>0$, so this condition varies temperature and selection together and cannot separate them mechanistically; we ran no intermediate-temperature sweep. And the selector is deliberately the cheap one---first non-empty result---not majority vote or reranking by \texttt{EXPLAIN} validity. The claim is therefore narrower than ``sampling does not help'': equal-budget sampling \emph{with a weak selector} is a worse use of the budget than grounded sequential correction. A competent selector would need the verification machinery this study is about, so the detector is load-bearing whether the budget is spent sequentially or in parallel.

% =============================================================================
\section{RQ3: Are We Measuring It Correctly?}
\label{sec:rq3}

\paragraph{n-gram overlap is not a bound in either direction.} Over the 1{,}917 retained result pairs, Google-BLEU correlates only moderately with set-based F1 ($r=0.71$; $\rho=0.59$), and disagrees asymmetrically. Usually it is \emph{optimistic}: BLEU exceeds set-F1 on $65.9\%$ of pairs and falls below on $3.1\%$ (mean $0.392$ vs.\ $0.229$), because serialised overlap credits rows that merely share tokens with rows that are wrong. In the other direction, $2.9\%$ of results are set-identical yet score below $1.0$ purely from row ordering, and set-identical results average $0.891$. So exact-match figures using BLEU $\geq 0.999$ as a proxy understate order-invariant accuracy for every model: naive retry's aggregate rises from $19.9\%$ to $22.6\%$ under direct set equality.

\paragraph{Against judged semantics it is pessimistic.} These findings are not in tension, since they compare BLEU to a stricter and a looser criterion. Against set membership it is too generous; against semantic equivalence too harsh, with $33.9\%$ of partial-BLEU queries judged fully correct. Overall $88.2\%$ of the 13{,}563 results are at least partially valid ($52.3\%$ correct), and recovered queries hold up at $83.3\%$ against $88.7\%$ for non-recovered, an expected gap since these are the queries the base model got wrong. No single scalar can adjudicate such systems; we report three invariance classes and leave AST or graph-pattern equivalence to future work \cite{ascoli2025etm,gao2024benchmark}.

\paragraph{The judge is optimistic against human labels.} Cohen's $\kappa = 0.63$ \cite{cohen1960kappa} between GPT-4o-mini and GPT-4o measures \emph{inter-model consistency}, not agreement with humans, and two judges sharing a training distribution can be consistently wrong. To test that directly, we labelled 122 corrected queries---every query for which we retain both a stored judge verdict and the executed result payloads---blind to those verdicts. Human and judge agree on $64.8\%$ of three-way decisions ($\kappa = 0.48$) and $82.8\%$ of valid/invalid decisions ($\kappa = 0.59$), both \emph{below} the inter-judge figures: LLM--LLM agreement overstates validity. The gap is directional---the judge calls $74.6\%$ at least partially correct against the human's $65.6\%$, a $9$-point optimism concentrated in over-use of PARTIAL---so the corpus-level $88.2\%$ should be read as an upper bound by roughly ten points. This bounds the weight the judge column in Table~\ref{tab:main} can carry but does not affect \S\ref{sec:rq2}, whose counterfactuals compare conditions scored identically. Appendix~\ref{app:judge} gives the full protocol for all three evaluators, together with the confusion matrix behind the $9$-point gap.

% =============================================================================
\section{Implications for Agent Design}
\label{sec:implications}

The ordering across the three counterfactuals is unambiguous: removing correction costs $3.1$--$12.3\%$, removing its schema grounding about one point of exact match, and reallocating the budget to sampling $10$--$11\%$. Three claims follow for any agent acting on a verifiable execution environment. \emph{Invest in the detector and the router, not the feedback generator}: where the environment returns a hard failure signal, most of the gain comes from noticing it and spending a bounded retry budget, and the marginal call that turns an error into eloquent guidance bought one point here. \emph{Sampling is not a substitute for grounded verification in low-entropy output spaces}: structured queries have few correct forms, so temperature moves mass off them, and any selector good enough to exploit diversity needs the verification machinery anyway. This is the conclusion budgeted verify-and-repair loops reach in code synthesis \cite{bhattarai2025arcs}. \emph{Evaluation must state its invariances}: order-sensitive overlap can over-report against set equivalence and under-report against semantics at once, leaving measured gains partly a metric artefact. The residual case for decomposition is real but modest---deterministic parsing saves a call, pre-execution validation saves round-trips, modules swap independently---and those are engineering properties, not accuracy gains.

% =============================================================================
\section{Conclusion}

We used LAST-CQ, a multi-agent framework for Text-to-Cypher that combines schema-aware validation with execution-grounded correction, as an instrument to ablate the agentic-refinement and order-invariant evaluation of agentic structured-query generation. It recovers $91.7\%$ of single-pass failures over 2{,}471 queries and six backbones, and three counterfactuals localise that gain to detection and orchestration.

In future work, we aim to expand the evaluation metrics using AST canonicalization and tree-matching, SMT-based checks, and validate semantics beyond surface results. Our interests include fine-tuning LLM models \cite{cazzaro2026achieving} and comparing them on the same benchmark dataset. Furthermore, latency and cost impact at large scale across multiple graph database engines represent another dimension to examine.

% =============================================================================
\section*{Limitations}

\paragraph{Single benchmark and engine.} All results come from the executable subset of one Neo4j benchmark. It is the only publicly available Text-to-Cypher benchmark with live databases for execution-based evaluation, but generalisation to Memgraph, to other property-graph engines, or to schemas outside these 16 domains is untested. \citet{gusarov2025multiagent} evaluate on CypherBench and Memgraph, so no direct numerical comparison with their system is possible. The counterfactual design itself is engine-agnostic, since it needs only a validator, an executor and a correction step, so porting it to a second engine and benchmark is the natural next experiment rather than a redesign.

\paragraph{Metric coverage is asymmetric.} Per-query result payloads were retained for the 1{,}917-query correction subset and a 214-query paired subset, but not for the full corpus, so set-based precision, recall, F1 and Jaccard are reported on those subsets rather than on all 2{,}471 queries per model. Full-corpus order-invariant evidence rests on exact match and judge accuracy. The paired head-to-head covers two of six backbones.

\paragraph{The comparison set is conditioned on our own failures.} The 1{,}917 queries in \S\ref{sec:naive} are those LAST-CQ chose to correct. Naive retry is never evaluated on queries LAST-CQ answered first-pass, so the comparison measures relative correction quality, not relative end-to-end system quality.

\paragraph{Substituted models.} Gemini 1.5 Flash and Pro were retired before the naive-retry experiment; both arms of that comparison use Gemini 2.5 Flash. Comparisons are always within-model, and excluding those two models does not change the conclusion, but the absolute numbers for those rows are not comparable to Table~\ref{tab:main}.

\paragraph{No-refinement is a counterfactual.} It re-scores an existing run rather than executing an alternative system, and therefore upper-bounds refinement's contribution; the single-pass Baseline gives the lower bound of $+3.1\%$, and we report both.

\paragraph{Equivalence is bounded, not demonstrated at arbitrary precision.} The schema-grounding null is established as equivalence within $\pm 0.075$ set-F1 on 214 paired queries across two backbones. A real advantage smaller than that cannot be excluded, and the paired evidence does not cover the other four backbones.

\paragraph{Best-of-3 confounds temperature with selection.} $T{=}0.7$ is required for diversity but differs from the $T{=}0$ baseline, and the selector is first-non-empty rather than a signal-using reranker. The claim is about equal-budget sampling as usually configured, not about sampling in principle.

\paragraph{Human calibration is single-annotator, and the judge has no baseline arm.} The 122-query calibration in \S\ref{sec:rq3} was labelled by one author with no second annotator, so it has no inter-human agreement figure of its own, and it covers corrected queries on two backbones rather than a corpus-wide stratified sample. It establishes the \emph{direction} and rough size of the judge's optimism, not a corrected corpus-level number. Separately, the judge was run only on LAST-CQ+DB outputs, so it cannot establish that the loop improves semantic correctness relative to single-pass generation, only that its outputs are largely valid.

\paragraph{Empty references interact with the metric convention.} $12.0\%$ of the corrected queries have empty reference results, which the benchmark's GLEU convention scores as $1.0$ for any non-empty output. The relaxation pass can therefore convert a correct empty answer into a scored-as-perfect wrong one. We quantify this in \S\ref{sec:naive} and report the exclusion analysis, but a benchmark with fewer empty references would test the loop more sharply.

\paragraph{Other.} One model (DeepSeek-Chat) shows a $1.3\%$ GLEU decrease under the full pipeline. Only DeepSeek-V3 publishes a parameter count; for two further backbones we can cite third-party estimates and for the remaining three no figure is available at all, so the small/medium/large groupings reflect vendor positioning and relative pricing rather than measured size (Appendix~\ref{app:models}). A systematic sweep over $N$ and $M$ was not performed; $N{=}2$ was chosen from the observed cycle distribution and $M{=}1$ is fixed by the implementation as a latch, so larger $M$ was never measured. We also did not ablate the individual hints in $H_{\text{relax}}$, run a blind-retry arm with routing but no feedback content, or record per-call token counts; all three would sharpen the attribution and are the natural next experiments.

% =============================================================================
\section*{Ethics Statement}

This work uses publicly available benchmarks, public demonstration databases and commercial LLM APIs. No private or sensitive data is processed. A system that turns natural language into executable database queries carries a risk of unintended data exposure if deployed against sensitive graphs; query-level access controls and result-scope limits should be enforced in production. We follow the ACL Code of Ethics.

\bibliography{custom}

\appendix

\section{Model Configurations}
\label{app:models}

We evaluate six foundation LLM backbones across three parameter scales (Table~\ref{tab:models}).
All models are used with temperature $T=0$ for deterministic outputs.
No model fine-tuning is performed; all generation relies on in-context prompting.
The model set extends previous work \cite{ozsoy2025text2cypher} with Claude Sonnet 4 and DeepSeek-Chat.

Parameter counts are disclosed for only one of the six backbones. DeepSeek-V3 publishes 671B total parameters with 37B active per token \cite{deepseek2024v3}. For GPT-4o and GPT-4o-mini we quote the third-party estimates tabulated by \citet{benabacha2025medec}, who label them as estimates rather than vendor figures. Google publishes no parameter count for either Gemini 1.5 model, and Anthropic publishes none for Claude Sonnet 4; we are not aware of a citable estimate for any of the three and therefore report them as undisclosed rather than guess. The scale grouping is consequently an ordering by vendor positioning and relative pricing, used only to keep model order consistent across tables.

\begin{table}[t]
\centering
\small
\begin{tabular}{llc}
\toprule
\textbf{Category} & \textbf{Model} & \textbf{Parameters} \\
\midrule
\multirow{2}{*}{Small} & GPT-4o-mini & $\sim$8B$^{\ast}$ \\
      & Gemini 1.5 Flash & undisclosed \\
\midrule
\multirow{2}{*}{Medium} & Claude Sonnet 4 & undisclosed \\
       & Gemini 1.5 Pro & undisclosed \\
\midrule
\multirow{2}{*}{Large} & GPT-4o & $\sim$200B$^{\ast}$ \\
      & DeepSeek-Chat & 671B (MoE) \\
\bottomrule
\end{tabular}
\caption{Evaluated backbones. Only DeepSeek-V3 has a vendor-published parameter count. $^{\ast}$Third-party estimate reported by \citet{benabacha2025medec}, not an official figure. The closed models otherwise disclose nothing, so the scale groupings reflect vendor positioning and relative pricing, not measured size.}
\label{tab:models}
\end{table}

\section{Judge and Annotation Protocol}
\label{app:judge}

Three evaluators appear in this paper at three different coverages. We describe each, since the differences between them bear on how much weight the judged numbers can carry.

\paragraph{Primary judge.} GPT-4o-mini at $T=0$, capped at 10 output tokens, rates each result on a three-point scale: CORRECT if the generated results fully answer the question equivalently to the ground truth, PARTIAL if they are related but incomplete, a superset, or slightly different, and INCORRECT if they fail to answer the question or return unrelated information. Each call receives the question, the ground-truth Cypher, the generated Cypher, and both serialised result payloads truncated to 500 characters. The judge sees results, not just queries, so it evaluates execution outcomes rather than surface query form.

\paragraph{Coverage, and what it excludes.} The judge was run over all $2{,}471 \times 6 = 14{,}826$ model--query pairs, of which $13{,}563$ returned a parseable verdict: per model, $2{,}164$ (GPT-4o-mini), $2{,}170$ (Gemini Flash), $2{,}265$ (Gemini Pro), $2{,}315$ (DeepSeek), $2{,}319$ (GPT-4o) and $2{,}330$ (Claude), i.e.\ $87.6$--$94.3\%$ coverage. The remainder are pairs for which no comparable payload was available. The semantic-accuracy column of Table~\ref{tab:main} therefore describes queries that executed and returned something to compare, which is a mildly optimistic base, and is one reason we treat that column as descriptive rather than as an ablation metric.

\paragraph{Second judge.} GPT-4o, same prompt and rubric, on 500 rows sampled with seed 42 and stratified over the three BLEU buckets used throughout ($166$ exact, $166$ high, $168$ partial). Raw agreement with GPT-4o-mini is $81.6\%$, Cohen's $\kappa = 0.63$. This measures inter-model consistency only.

\paragraph{Human annotation.} The 122 human-labelled queries are a census, not a sample: they are every query for which both a stored judge verdict and the executed result payloads survive, which restricts them to two backbones (Gemini-Pro, $67$; DeepSeek, $55$). Rows were shuffled with seed 42 and presented with the judge verdict withheld in a separate key file, so the annotator---one of the authors---labelled blind and to the same three-point rubric. Table~\ref{tab:confusion} gives the resulting confusion matrix. The judge's optimism is concentrated in a single cell: $15$ of the $42$ queries the annotator called INCORRECT were rated PARTIAL by the judge, against only $1$ rated CORRECT. The disagreement is therefore not noise around the boundary of correctness but a systematic reluctance to reject.

\begin{table}[t]
\centering\small
\begin{tabular}{lccc}
\toprule
& \multicolumn{3}{c}{\textbf{Judge}} \\
\cmidrule(lr){2-4}
\textbf{Human} & \textbf{CORR.} & \textbf{PART.} & \textbf{INCORR.} \\
\midrule
CORRECT   & 36 & 18 & 3  \\
PARTIAL   & 4  & 17 & 2  \\
INCORRECT & 1  & 15 & 26 \\
\bottomrule
\end{tabular}
\caption{Human versus GPT-4o-mini verdicts on the 122 blind-labelled queries. Three-way agreement $64.8\%$ ($\kappa=0.48$); collapsing CORRECT and PARTIAL to valid gives $82.8\%$ ($\kappa=0.59$).}
\label{tab:confusion}
\end{table}

\paragraph{What this protocol does not establish.} The three evaluators cover different populations, chosen by what data survived rather than by design: the primary judge covers most of the corpus, the second judge a stratified 500, and the human annotator a two-backbone census of corrected queries, which are harder than average. The human figures therefore fix the direction and rough magnitude of the judge's optimism, not a corrected corpus-level accuracy, and none of the three was run on baseline outputs.

\section{Additional Analyses}
\label{app:extra}

\begin{figure}[t]
\centering
\includegraphics[width=\columnwidth]{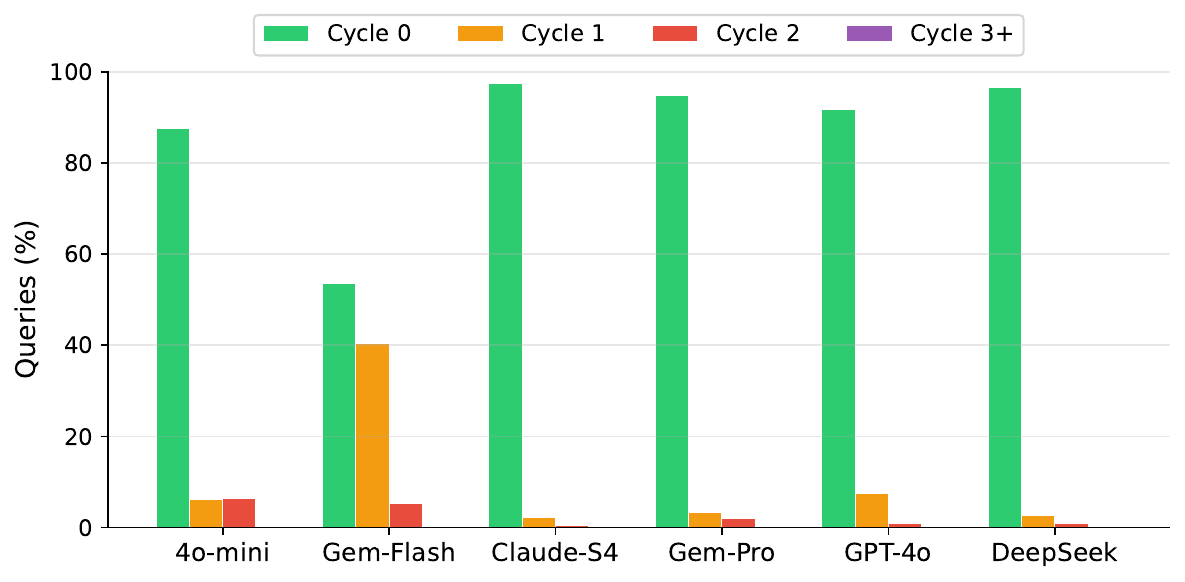}
\caption{Validation cycle distribution per model.}
\label{fig:cycles}
\end{figure}

\paragraph{Validation cycles.} Fewer than $13\%$ of queries undergo any validation cycle for most models (Figure~\ref{fig:cycles}); Gemini Flash is the exception at roughly $50\%$, tracking its baseline failure rate. Of the 1{,}917 queries entering the loop, $80\%$ resolve in one cycle and $20\%$ require two. Quality degrades gracefully with cycle count: one-cycle corrections reach mean GLEU $0.36$--$0.55$, comparable to zero-cycle queries ($0.43$--$0.54$), while two-or-more-cycle queries range from $0.14$ (Gemini Flash) to $0.54$ (GPT-4o).

\paragraph{Efficiency detail.} Mean validation cycles per query are $0.19$, $0.50$, $0.03$, $0.07$, $0.09$ and $0.04$ for GPT-4o-mini, Gemini Flash, Claude, Gemini Pro, GPT-4o and DeepSeek respectively, with $87.6\%$, $53.5\%$, $97.5\%$, $94.7\%$, $91.7\%$ and $96.6\%$ of queries resolved without any correction.

\paragraph{Bootstrap confidence intervals.} Table~\ref{tab:main} values carry $95\%$ bootstrap CIs of $\pm 0.015$ GLEU on average (e.g.\ GPT-4o: $0.5421$, $[0.5284, 0.5559]$). Cross-model differences are robust; per-model Baseline-to-LAST-CQ+DB improvements (e.g.\ Claude, $+0.0078$) fall within overlapping intervals, consistent with our framing that recovery, not incremental gain on already-working queries, is the effect of interest.

\paragraph{Result-quality distribution.} Across models, $23\%$ of results reproduce the reference exactly under GLEU, $18\%$ score above $0.5$, $58\%$ below, and under $1\%$ share no overlap. Section~\ref{sec:rq3} shows why the middle band is dominated by metric behaviour rather than by semantic failure.

\paragraph{Error-type recovery and the DeepSeek regression.} Among 239 corrected queries with classifiable errors, schema violations are recovered in $88\%$ of 75 cases and runtime errors in $49\%$ of 119, while pure syntax errors are never recovered ($0/31$) and type mismatches almost never ($8\%$ of 13). Representative successful corrections rewrite SQL-style \texttt{FROM} clauses into \texttt{MATCH} traversals and string-based date comparisons into \texttt{date()} calls. The DeepSeek GLEU regression has a separate cause: LAST-CQ's normalised schema alters property ordering and label casing relative to the raw schema used at baseline, causing 576 of 593 affected queries to generate \emph{different} rather than worse Cypher, with only 17 cases of genuine correction drift. DeepSeek's recovery rate remains $93\%$.

\paragraph{Empty-reference queries.} $230$ of the $1{,}917$ corrected queries have an empty reference result. Under naive retry $54$ of them return non-empty output; in the 214-query paired subset, $3$ of $33$ do so under LAST-CQ. Both are scored $1.0$ by the benchmark's GLEU convention despite being wrong.

\paragraph{Schema parser coverage.} Deterministic parsers handle $100\%$ of benchmark schemas using two of the eight supported formats ($97.9\%$ and $2.1\%$); the LLM fallback never fires. Non-standard deployment schemas could still require it.

\paragraph{Set-metric definitions.} For canonicalised row sets $A$ (generated) and $B$ (reference), we report $\mathrm{P} = |A \cap B|/|A|$, $\mathrm{R} = |A \cap B|/|B|$, their harmonic mean, $\mathrm{J} = |A \cap B|/|A \cup B|$, and set equality $\mathbf{1}[A = B]$. Each row dict is reduced to a canonical sorted key--value tuple before set construction, so both row order and key order are discarded. When exactly one set is empty all measures are $0$; when both are empty all are $1$.

\end{document}